\documentclass[a4paper,conference]{IEEEtran}

\usepackage{amsmath}

\usepackage{url}
\usepackage{hyperref}

\usepackage[pdftex]{graphicx}
\graphicspath{/img}
\DeclareGraphicsExtensions{.jpeg,.png}

\usepackage{booktabs}
\usepackage{xspace}
\makeatletter
\DeclareRobustCommand\onedot{\futurelet\@let@token\@onedot}
\def\@onedot{\ifx\@let@token.\else.\null\fi\xspace}

\def\ie{\emph{i.e}\onedot}

\def\wrt{w.r.t\onedot} 

\makeatother

\newcommand{\sei}{\textsc{I-Split}\xspace}
\newcommand{\cui}{\textsc{CUI}\xspace}
\newcommand{\gradcam}{Grad-CAM\xspace}

\usepackage[dvipsnames]{xcolor}
\usepackage{xcolor}
\usepackage{soul}
\setulcolor{Red}

\newcommand{\rev}[1]{\textcolor{Magenta}{#1}}

\newcommand{\federico}[1]{\textcolor{OliveGreen}{[Federico: #1]}}

\begin{document}

\title{\sei{}: Deep Network Interpretability for Split Computing}

\author{
\IEEEauthorblockN{Federico Cunico, Luigi Capogrosso, Francesco Setti, Damiano Carra, Franco Fummi, Marco Cristani}
\IEEEauthorblockA{\textit{Department of Computer Science, University of Verona}}
{\tt name.surname@univr.it}
}

\maketitle

\begin{abstract}
This work makes a substantial step in the field of split computing, \ie{}, how to split a deep neural network to host its early part on an embedded device and the rest on a server. So far, potential split locations have been identified exploiting uniquely architectural aspects, \ie{}, based on the layer sizes. Under this paradigm, the efficacy of the split in terms of accuracy can be evaluated only after having performed the split and retrained the entire pipeline, making an exhaustive evaluation of all the plausible splitting points prohibitive in terms of time.
Here we show that not only the architecture of the layers does matter, but the \emph{importance} of the neurons contained therein too. A neuron is important if its gradient with respect to the correct class decision is high. It follows that a split should be applied right after a layer with a high density of important neurons, in order to preserve the information flowing until then. Upon this idea, we propose  \emph{Interpretable Split} (\sei{}): a procedure that identifies the most suitable splitting points by providing a reliable prediction on how well this split will perform in terms of classification accuracy, \emph{beforehand} of its effective implementation. As a further major contribution of \sei{}, we show that the best choice for the splitting point on a multiclass categorization problem depends also on which specific classes the network has to deal with.
Exhaustive experiments have been carried out on two networks, VGG16 and ResNet-50, and three datasets, Tiny-Imagenet-200, notMNIST, and Chest X-Ray Pneumonia.
The source code is available at \url{https://github.com/vips4/I-Split}.
\end{abstract}

\section{Introduction}
\label{cha:intro}

In the last decade, Deep Neural Networks (DNNs) achieved state-of-the-art performance in a broad range of problems. However, DNN computational requirements preclude their \emph{device-only} deployment on most of the resource-constraint systems, such as mobile phones~\cite{matsubara2021split}. 
The opposite \emph{cloud-only} paradigm transfers the sensory data to a server~\cite{tsai2010service}, which processes them and sends the output back to the device. In this case, the high data transfer time and possible server congestion are obvious downsides. 
Between the two extremes lies the \emph{fog computing}  paradigm~\cite{singh2019fog}, where computation is distributed over a large number of low-power devices, and communication is routed over the internet backbone.

\begin{figure}[t]
\begin{center}
\includegraphics[width=.95\linewidth]{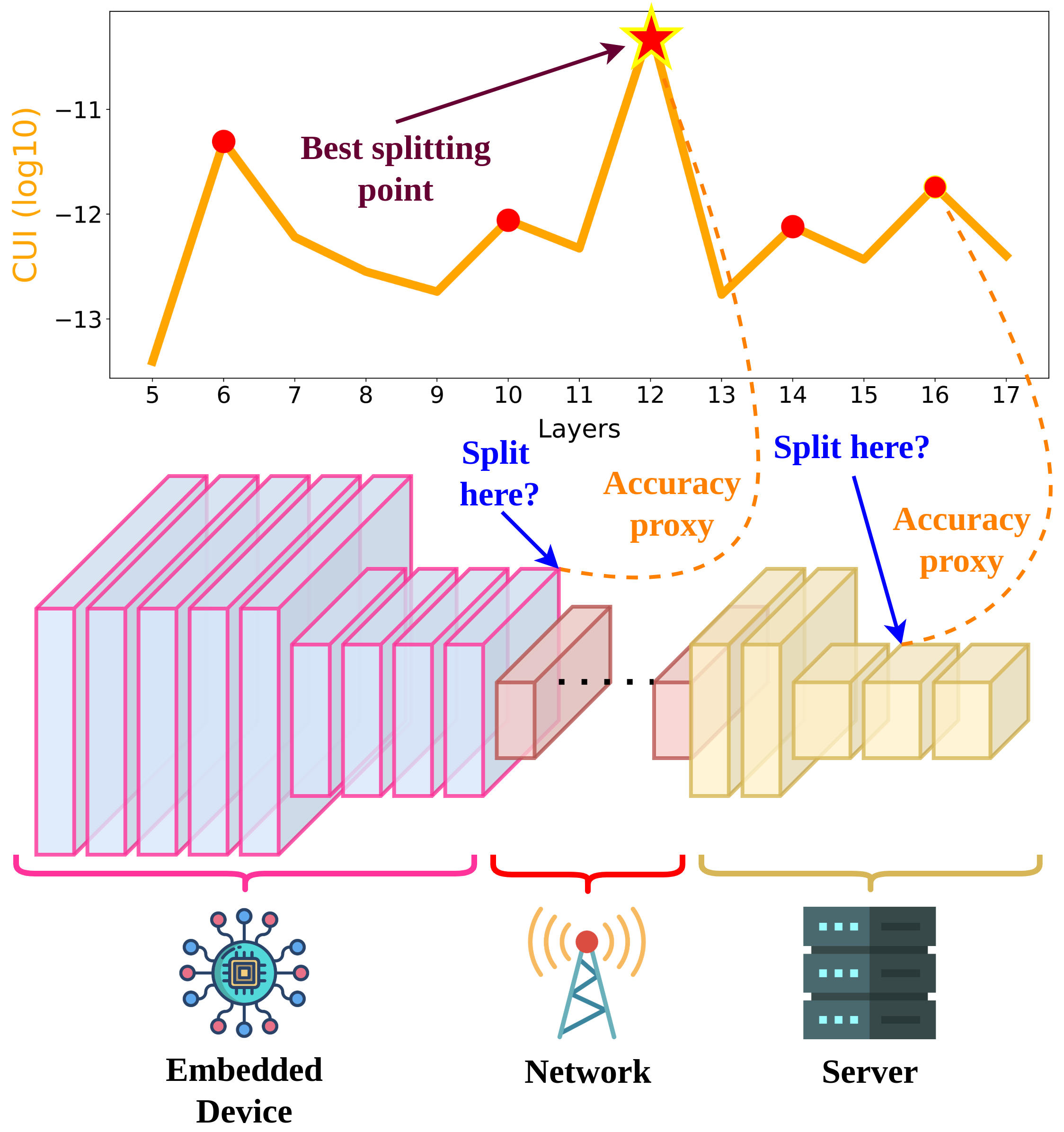}
\end{center}
\caption{\sei{} teaser: the DNN architecture is divided into a ``head'' part running on the edge device and a ``tail'' part running on a server machine. Intermediate data representation is sent through a network connection. The optimal splitting point is determined by analyzing the value of the \textsc{CU}mulated \textsc{I}mportance curve (\cui{}), which indicates the importance of each layer in predicting the correct class. The higher the \cui{}, the higher the amount of accuracy of the original model preserved when splitting at that layer.
\vspace{-1.8em}}
\label{fig:Schema}
\end{figure}

As a compromise between the device-only and the cloud-only approaches or to drive the design of a fog computing architecture, the \emph{split computing} (SC) framework~\cite{shao2020bottlenet++,sbai2021cut,matsubara2022supervised} proposes to divide a DNN model into a ``head'' and a ``tail'', deployed at the edge device and the server, respectively. %
Using SC, the computational load is distributed across the two platforms, exploiting the computational power of both devices at the cost of transferring data on a network connection. Data transfer effort is mitigated by the injection of \emph{bottlenecks}, usually implemented as autoencoders~\cite{matsubara2019distilled}, achieving in-model compression. %
It is worth noting that SC is even gaining more and more relevance in the 5G Telecom scenario, which is intrinsically based on the edge/cloud paradigm, thus requiring SC by design~\cite{du2019definition}.

Setting up a SC system requires determining the optimal splitting point: so far, this choice was driven primarily by architectural considerations such as memory capabilities of the edge device, desired transmission rate between head and tail, and size of the layers~\cite{sbai2021cut}. %
In all of these cases, deciding a splitting point is blind to a crucial feature a deep network has to preserve: its classification accuracy. For example, two consecutive layers of the same size are equally likely to be chosen as splitting points, and the accuracy of the final system has to be verified only after retraining the head, bottleneck, and tail. This brings to a cumbersome trial-and-error cycle.

In this paper, we propose a fast procedure to select the split location for a generic DNN architecture that, for the first time, is predictive of the accuracy that the system will have once retrained (see Fig. \ref{fig:Schema}), thus eliminating the need for multiple trials and error retraining sessions of the system. The procedure is dubbed \sei{}, where ``\textsc{I}'' stands for interpretability. \sei{} builds upon the concept of \emph{importance} or saliency of a neuron~\cite{selvaraju2017grad}, which is related to the gradient it possesses with respect to the decision towards the correct class, for specific input. %
Importance is exploited with success in the \gradcam{} approach~\cite{selvaraju2017grad}: \gradcam{} creates an input neuron saliency map that indicates which parts of an input image are more important for deciding a specific class. In particular, the \gradcam{} approach has been proved to be strongly dependent on the given trained model on which it runs (it passes the ``sanity check'' test~\cite{adebayo2018sanity}), while other approaches do not, making it perfectly suited to our purposes.

\sei{} exploits \gradcam{} by creating, for a given image, multiple saliency maps, one for each layer of the network, which we rename as \emph{importance maps}.
Each layer-based importance map can be accumulated in a single value, accounting for how many important neurons it is formed by. Therefore, a single image gives rise to multiple \emph{\textsc{CU}mulated \textsc{I}mportance} (\cui{}) values, one for each layer, that can be rearranged into a \cui{} curve.  Multiple validation images create multiple \cui{} curves, that summed together do create a statistic of how much a layer is, in general, decisive for the right class. A layer that exhibits a high \cui{} value needs to be preserved, \ie{}, the bottleneck should be injected right after this layer. Higher \cui{} values are predictors of high accuracy, and the ranking over \cui{} allows to easily select the optimal splitting point.

Several are the advantages of \sei{}. The process is computationally efficient: the evaluation of the \cui{} values for $N$ potential splitting points requires one step of backpropagation for each image of a given validation set, instead of $N$ complete retraining sessions; in practice, for 10 potential splitting points to evaluate, \sei{} requires 150 minutes on a VGG network, instead of 6 hours needed to for the retraining.
Moreover, we are able to discriminate, among layers of the same size which one is best suitable as the splitting point.  
Finally, and most interestingly, our approach shows that optimal splitting points are conditioned on the specific classes that the network is expected to process: indeed, specific classes trigger specific neurons, which in turn highlight layers which are possibly different. This provides the fresh-new concept of \emph{class-dependent split decision}, which allows us to adapt the splitting point depending on the classes taken into account.

The experiments are conducted on two popular network architectures (VGG16~\cite{simonyan2014very} and ResNet-50~\cite{he2016deep}), and three classification benchmarks (Tiny-Imagenet-200~\cite{le2015tiny}, notMNIST~\cite{notMNIST}, and Chest X-Ray Pneumonia~\cite{chest_xray}), confirming each of the claims discussed above and surpassing previous empirical split strategies as in~\cite{sbai2021cut}. 

In summary, the contributions of this paper are:
\begin{itemize}
\item \sei{}, a principled and fast way to individuate splitting points in a DNN, together with an indication of how the networks split and those points will perform in terms of classification accuracy;
\item With \sei{}, we show that specific classes bring different splitting points.
\end{itemize}

\begin{figure*}[t]
\begin{center}
\includegraphics[width=\linewidth]{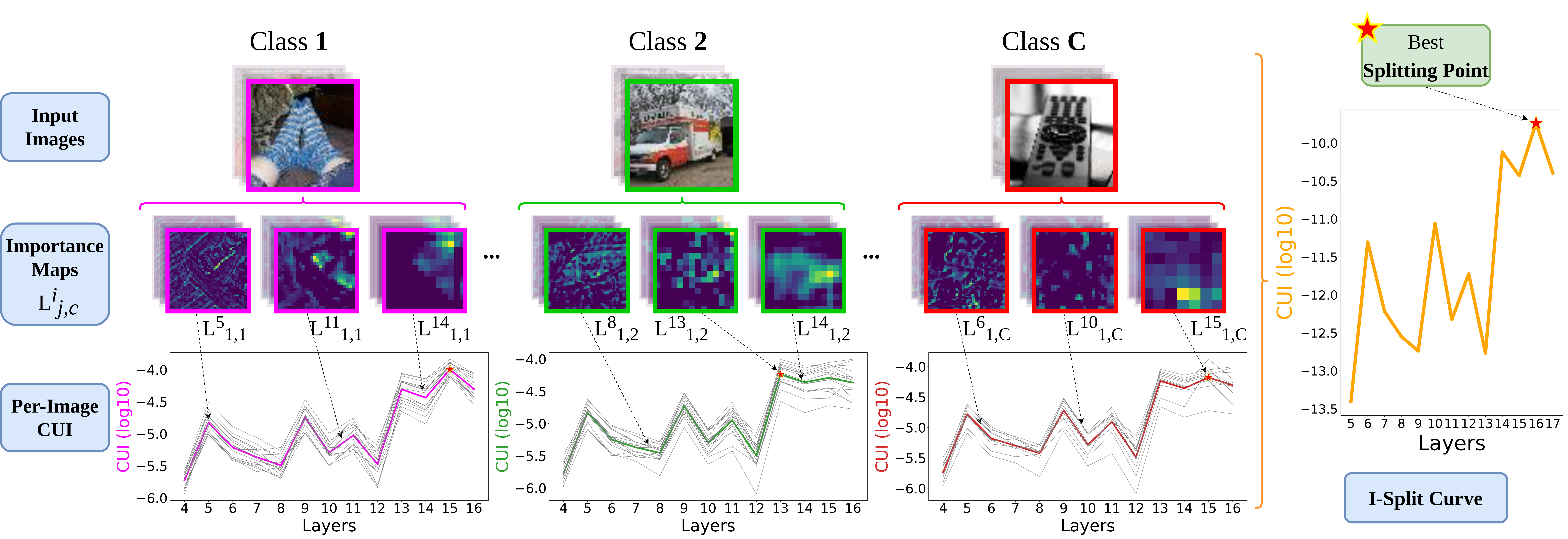}
\end{center}
\caption{Overview of our \sei{} framework. The input images are fed into a neural network to extract high-resolution importance maps using the \gradcam{} algorithm at each layer. Then, we average over all the image pixels of each map to produce per-image \cui{} curves. Finally, all curves are fused to generate the general \cui{} curve. The best splitting point for the network is the global maximum of the \cui{} curve.}
\label{fig:Figure_2}
\end{figure*}

\section{Related Work}
\label{cha:related}

This section provides an overview of existing state-of-the-art split computing approaches along with a discussion of different interpretation methods for neural networks. Interested readers can refer to surveys~\cite{matsubara2021split} and~\cite{olah2018building}.

\subsection{Split Computing Methods}
A typical split computing scenario is discussed in~\cite{eshratifar2019jointdnn}, where the authors show that neither cloud-only nor device-only approaches are optimal, and a split configuration is the ideal solution.
In~\cite{kang2017neurosurgeon}, it is shown that the early layers of a DNN are the most suitable for a split since they optimize latency and energy consumption. Furthermore, latency can be reduced by either quantization~\cite{li2018auto} or lossy compression before data transmission~\cite{choi2018deep,cohen2020lightweight}. The use of autoencoders as a way to further diminish the data to be transferred is discussed in~\cite{matsubara2019distilled,eshratifar2019bottlenet,jankowski2020joint}. All these approaches have an impact on the overall accuracy, therefore some works explore the trade-off between latency and accuracy~\cite{sbai2021cut,hu2020fast,yao2020deep}. The impact on the accuracy is also mitigated by additional techniques, such as knowledge distillation~\cite{matsubara2021supervised,matsubara2021neural}. To find candidate splitting points, the most used approach is the ``Cut, Distil and Encode'' (CDE)~\cite{sbai2021cut}. It states that candidate split locations are where the size of the DNN layers decreases: the rationale is that compressing information by autoencoders, where compression would still occur due to the shrinking of the architecture, certainly seems reasonable.

Notably, all these approaches first identify a set of candidate splitting points, then they re-train the split models to find the one with the highest accuracy. In this paper, we present a way to individuate the best splitting point before any retraining.  

\subsection{Interpretability Methods}
In deep learning, \emph{interpretability} refers to the ``intrinsic properties of a deep model measuring to which degree the inference result of the model is predictable or understandable to human beings''~\cite{li2021interpretable}. %
It can be distinguished into intrinsic and post-hoc~\cite{stiglic2020interpretability}: intrinsic interpretability applies to models that have a simple analytical form such as short decision trees or  sparse linear models~\cite{carvalho2019machine};  post-hoc interpretability refers to the application of interpretation methods once a model has been trained~\cite{peake2018explanation}. %
In this paper, we focus on post-hoc methodologies, and in particular, on \emph{gradient-based techniques}~\cite{adebayo2018sanity}, also known as ``saliency-based'' approaches. %
These techniques are based on the computation of the partial derivative of the prediction $y^c$ for the class $c$ with respect to image input $x$, that is, $\partial y^c/\partial x$~\cite{baehrens2010explain}. %
The derivative indicates how much a change in each input feature would change $y^c$; simple but effective, an approach that exploits this principle is the Gradients technique~\cite{adebayo2018sanity}. The element-wise product of the input and the gradient represents another popular method~\cite{shrikumar2016not}: the idea is that the gradient communicates the importance of a dimension, and the input indicates how strongly this dimension is expressed in the input image.
In the same paper, it is proposed the DeepLift approach, which explains the difference in the saliency map from some reference maps in terms of the difference between the input image from some reference input. The Integrated Gradients method~\cite{sundararajan2017axiomatic} addresses the \emph{sensitivity} and the \emph{implementation invariance} properties of saliency maps. The sensitivity states that if a single feature of an input is responsible for deciding a class, it has to be highlighted. The implementation invariance means that saliency maps should be the same independently on the models if the models have the same input/output pairs. %
Guided Backpropagation~\cite{springenberg2014striving} focuses on the quality of the saliency maps, combining vanilla backpropagation and DeconvNets to obtain saliency maps with a high level of detail. %
The \gradcam{} technique, at the basis of our approach, operates a downstream summation of class-specific coefficients multiplied by activation maps~\cite{selvaraju2017grad}. %
Notably, the \gradcam{} approach has been proved to pass the ``sanity check'' for saliency-based interpretability approaches ~\cite{adebayo2018sanity}, together with the Gradients technique. In practice, this check ensures that the saliency maps provided as output depend on the specific model instance taken into account, while other saliency-based approaches do not. Since our split strategy operates on a given specific model by finding split points as summations over these saliency maps, this kind of dependence is crucial and desirable.



\section{Methodology}
\label{cha:method}

The final goal of \sei{} is to individuate and rank potential splitting points in a generic DNN architecture so that the accuracy values obtained by the DNN, once split at those points, follow a similar ranking. The input of the approach is a network that has been trained beforehand on some training images.
The overview of the approach is shown in Fig.~\ref{fig:Figure_2}: each image in a pool of evaluation images, we compute importance maps associated with each layer of the network instead of being associated with input images only. These importance maps are then aggregated to form per-image \cui{} curves (the light-colored curves). Then, these curves are averaged together over the classes giving the general \cui{} curve on the right in yellow. Each point of the \cui{} curve is found to be proportional to the accuracy that the network will have, on the same data, once the split has been applied and the network re-trained. The highest point in the \cui{} curve identifies the location where the split will provide the best performing system.  

In the following, we will give the mathematical details to generate the \cui{} curve. Then we will detail how to compute the split in correspondence to a given location (Sec.~\ref{sec:split_app}) and how to re-train the network (Sec.~\ref{sec:re-train}).  


\subsection{\cui{} Curve Computation}
We assume our neural network model is pre-trained on a given training set. The \cui{} curve is computed on some validation set composed of $C$ classes, each formed by $N_C$ images. For each $c$-th class, we do as follows.

For a given $i$-th layer of the neural network, $i=1,\ldots,I$, we extract the class-discriminative activation map $L^{i}_{j,c}$ for each $j$-th image belonging to class $c \in C$. 
For this sake, we start by computing the feature map importance coefficient of \gradcam{} $\alpha{}_{i,j}^{c}$: 
\begin{equation}
\alpha{}_{i,j}^{c}=\frac{1}{z}\sum_{n}^{}\sum_{m}^{}\frac{\partial y^{c}}{\partial F_{n,m}^{i,j}}
\end{equation}
where $F^{i,j}\in R^{n \times m\times z}$ is the feature map of the convolutional layer $i$ for the image $j$.  

The weight $\alpha{}_{i,j}^{c}$ represents a partial linearization of the deep network downstream from $F$ and captures the value of the feature map of the $i$-th layer for a target class $c$. Then, we perform a weighted sum between the value just calculated and the feature maps $F^{i,j}$ of the chosen layer. Finally, \emph{ReLU} activation function is applied to reset the negative values of the gradient to zero obtaining the class activation map $L^{i}_{j,c}$ for a specific query image $j$:
\begin{equation}\label{eq:gradcam}
L^{i}_{j,c}=ReLU\left(\sum_{k=i}^{I}\alpha{}_{k,j}^{c}F^{k,j}\right)
\end{equation}
This represents the analog of the saliency map of the standard \gradcam{}~\cite{selvaraju2017grad}, which instead of being computed on the input image, is focused on the $i$-th layer, summing from the class activations of the networks back until $i$. 

Now we aim to obtain a single value for our class activation map $L^{i}_{j,c}$: thus, we simply sum over the dimensions of $F^{i,j}\in R^{n \times m\times z}$, obtaining the \emph{per-image} \cui{} values $\cui_{j,c}^{i}$. Ideally, computing these values for each $i$-th layer of the network gives a curve showing how the image has triggered the different layers of the network (see Fig.~\ref{fig:Figure_2}, the pool of curves below each class).
At this point, averaging over all the images of all the classes provides our final cumulated importance \cui{} curve (Fig.~\ref{fig:Figure_2} the yellow curve on the right), where the $i$-th point $\cui^i$ of the curve is a surrogate of the information conveyed through the $i$-th layer \emph{towards the decision for the correct class}, for all the classes into play. 
It is worth noting that, possibly, one may limit to compute \cui curve only on specific classes (or set of classes), obtaining \emph{per-class} curves $\cui^{i}_c$.

\subsection{Split Application} \label{sec:split_app}
Once the \cui{} curve has been computed, candidate splitting points can be individuated by \emph{(i)} choosing the layer which gives the highest peak or \emph{(ii)} selecting layers that give local \cui{} maxima, if other constraints than the accuracy have to be taken into account.

Let $T^{i}$ be the target layer for splitting the model at index $i$ and $T^{i+1}$ the subsequent layer. %
We divide the network into three main blocks: The head, running on the edge device, is composed of the first layers of the original DNN architecture, up to layer $T^i$; the \emph{bottleneck}, an under-complete autoencoder~\cite{pinaya2020autoencoders} that learns low-dimensional latent attributes which explain the input data; and the tail, from layer $T^{i+1}$ to the very end of the network, that is executed on the server-side. The encoder part of the bottleneck is deployed to the edge device, while the decoder is executed on the server-side. Encoders and decoders use both spatial and channel-wise reduction/restoration units, respectively. 

\subsection{Re-Training Strategy} \label{sec:re-train}
In order to train the entire model $M$, we first train our bottleneck, and then we fine-tune end-to-end the entire network.
We create an under-complete autoencoder which acts as a bottleneck inserted after $T^{i}$. This bottleneck should learn to replicate the input, which is the feature map in output at layer $T^{i}$. Therefore, given $\{\boldsymbol{I}_j, j=1,2,...,n\}$ as the $n$ number of training data, we train the sole bottleneck freezing the remaining network with the following loss:
\begin{equation}
\mathcal{L}_{AE} = \frac{1}{n} \sum_{j=1}^{n} || \Phi_{T^{i}}(\boldsymbol{I}_j) - \Psi(\Phi_{T^{i}}(\boldsymbol{I}_j); \boldsymbol{W}_{AE})||^2
\end{equation}
with $\Phi_{T^{i}}$ the model layers up to the $i$-th layer $T^{i}$ (target layer), $\Psi$ is the AE part of the model with weights $\boldsymbol{W}_{AE}$.

After training the bottleneck, we perform a fine-tuning of the network with the following loss function:
\begin{equation}
\mathcal{L}_{task} = \frac{1}{n} \sum_{j=1}^{n} || \Phi_M(\boldsymbol{I}_j;\boldsymbol{W}_M) - \hat{y}_j ||^2
\end{equation}
with $\Phi_M$ the full $M$'s DNN layers, $\boldsymbol{W}_{M}$ the weights of the model $M$ and $\hat{y}_j$ the ground truth label for the image $j$.

\section{Experimental Results}
\label{cha:experiments}

In this section, we show how \sei{} individuates split points by the \cui{} curve so that the associated \cui{} values are predictive of the classification accuracy when the network is split at those points. This hypothesis is evaluated on three benchmarks and two classification networks (Sec.~\ref{sec:split_analysis}). Moreover, we show that the \cui{} curve and the associated best split point (correspondent to the highest \cui{} peak) do depend on the classes on which we are focusing (Sec.~\ref{sec:split2class}). 

In all the experiments, the bottlenecks injected in the split networks are standard convolutional under-complete autoencoders with two convolutional layers with stride 2 for the encoder and two layers for the decoder. The number of filters depends on the desired compression rate: specifically, we keep constant the compression rate at 90\%, which means that the dimensionality of the encoded data is 10\% of the dimension of the splitting layer.

\subsection{Splitting Point Analysis} \label{sec:split_analysis}
\begin{figure}[t]
\begin{center}
\includegraphics[width=0.95\linewidth]{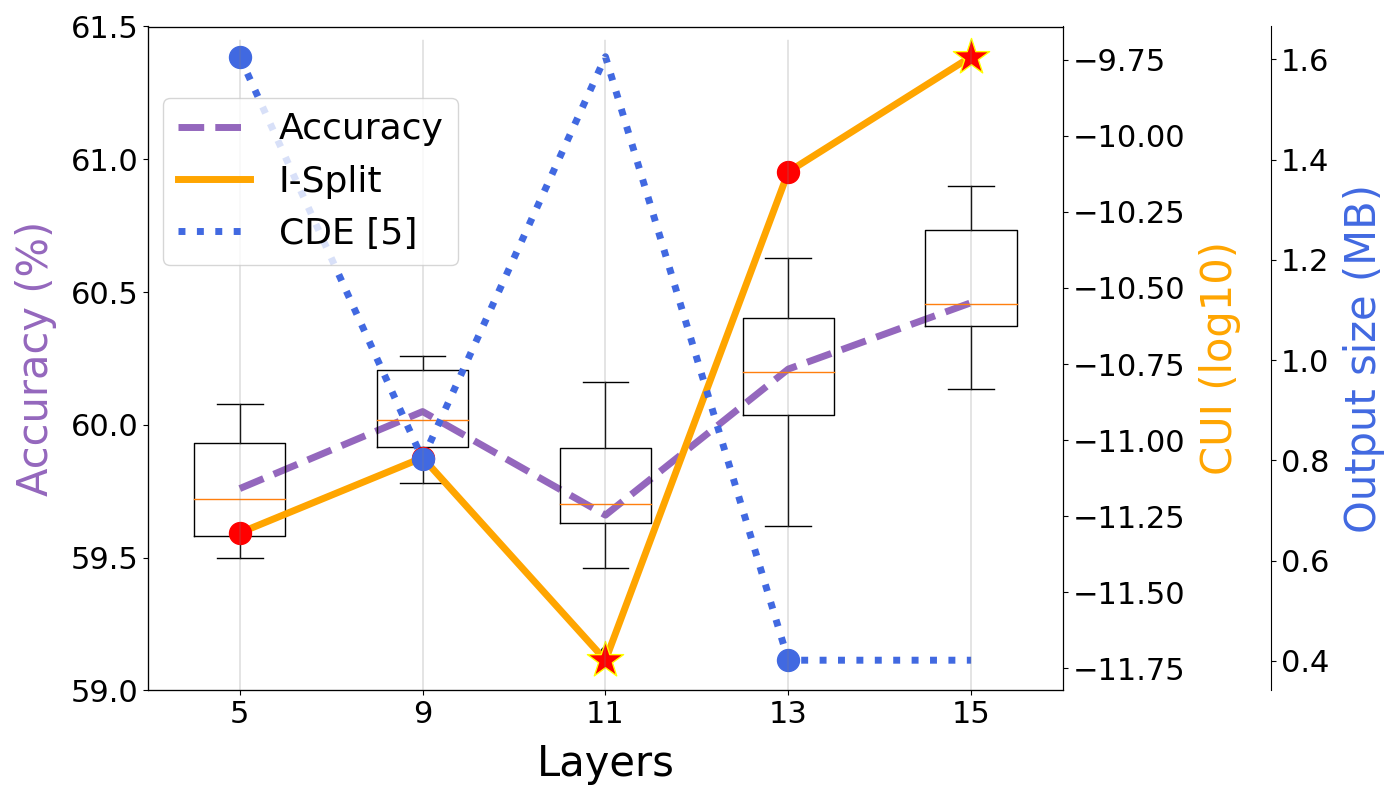}
\end{center}
\caption{Candidate splitting points identified by CDE and the \cui{} curve of \sei{} on the VGG16. In dashed purple, the accuracy values are obtained by splitting the network at that point, re-training it, and testing on a validation set. The ``star'' markers show the extra points that our method is able to identify, and not identified by the CDE approach.
\vspace{-.8em}}
\label{fig:CDE_and_SEI_and_accuracy_VGG16}
\end{figure}

Since our framework is agnostic with respect to the network architecture, we consider the VGG16 and the ResNet-50 networks, with most of the experiments run on the first one. Other than being very popular architectures, they are also interesting for SC: they exhibit a large number of parameters, a good depth in terms of the number of layers, with a dramatic variation in the layer dimensionality as depth increases.

The first experiment focuses on the Tiny-Imagenet-200 dataset~\cite{le2015tiny}, a compact version of ImageNet that comprises a subset of 200 classes, with a training set of 100.000 images, a validation, and a test set of 10.000 images each. All the images are 64$\times$64 pixels.

As a comparative approach, we consider CDE~\cite{sbai2021cut}, which individuates splitting points whereas there is a decrease in the size of the layers (see Sec.~\ref{cha:related} for more details). The multi-axis Fig.~\ref{fig:CDE_and_SEI_and_accuracy_VGG16} reports the CDE curve in blue, with the markers indicating the candidate split points. Our \sei{} is reported with the curve in solid yellow, with the red markers individuating local maxima of the \cui{} curve to simplify the visualization. Importantly, we report also the accuracy curve, in dashed purple, obtained by retraining the network after splitting it at the selected locations.  

Several facts do emerge:
\emph{(i)} Our \sei{} identifies all candidate splitting points output by CDE (namely layers $5$, $9$ and $13$, corresponding to \emph{block2\_pool}, \emph{block3\_pool}, and \emph{block4\_pool}, respectively): these are max-pooling layers in the VGG architecture, which are worth conveying more information-per-pixel since there is a local dimensionality reduction with limited loss of information. %
\emph{(ii)} \sei{} finds two additional points at layers $11$ and $15$, or \emph{block4\_conv2} and \emph{block5\_conv2}, that do not correspond to a decrease in the layer size. %
\emph{(iii)} Looking at the accuracy curve, it becomes apparent the significance of the \cui{} curve: all the candidate splitting points are proportional to the post-hoc accuracy obtained by the split network, while CDE does not exhibit such behavior at all. \emph{(iv)} While the absolute values of classification accuracy are relatively close to each other, we provided a statistical analysis generating 15 testing sets by randomly sampling 800 images from the validation set. %
Box plots in Fig.~\ref{fig:CDE_and_SEI_and_accuracy_VGG16} demonstrate how the distributions are well separated, and thus the relatively small improvement in accuracy is statistically relevant.

\begin{figure}[t]
\begin{center}
\includegraphics[width=.95\linewidth]{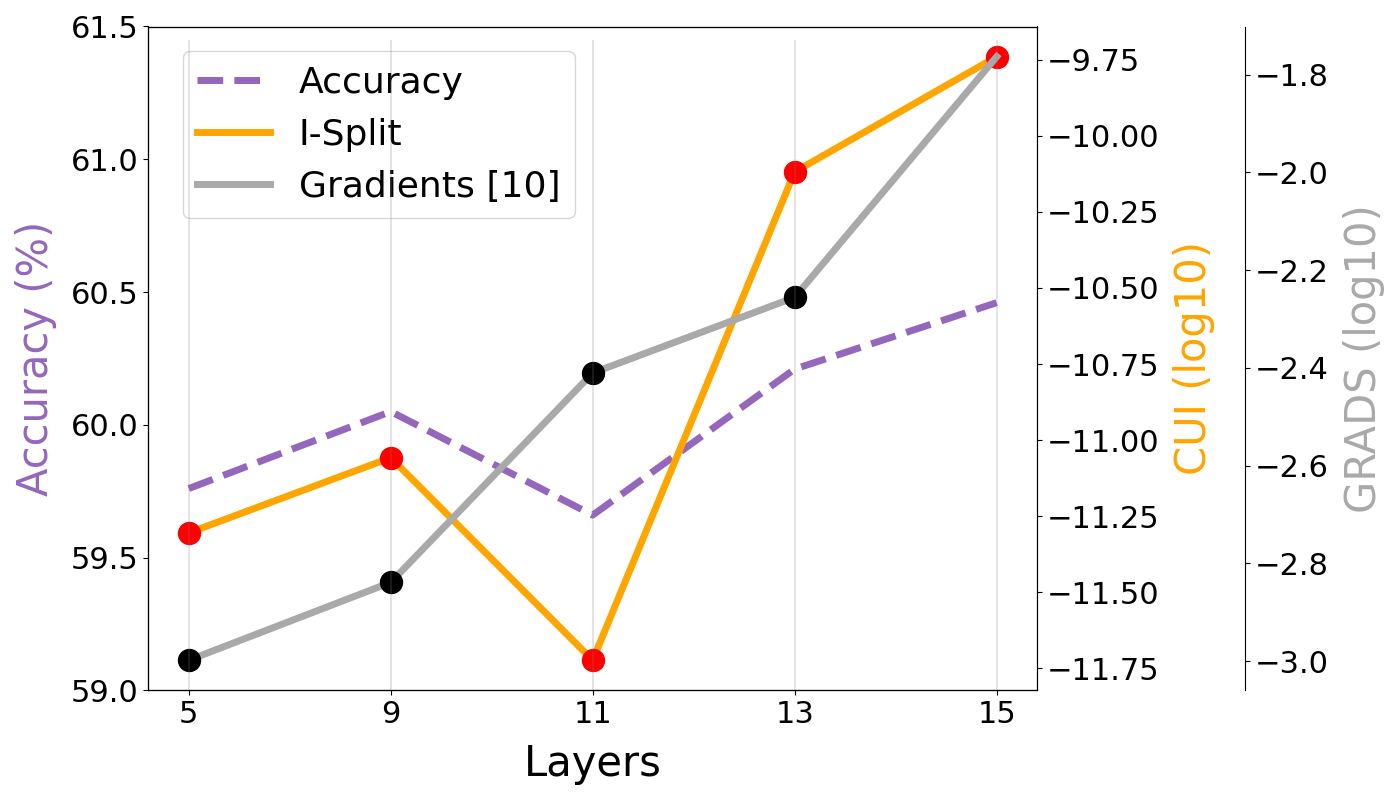}
\end{center}
\caption{The sensibility of our \sei{} method to the chosen interpretability approach: in gray the \cui{} using the \emph{Gradients} approach. In yellow, \gradcam{}. As visible, \gradcam{} gives a better curve in terms of proportionality \wrt{} the final class accuracy, while \emph{Gradients} is growing as soon as the layer is deeper.
\vspace{-1em}} 
\label{fig:SEI_grad}
\end{figure}
\sei{} strongly relies on the \gradcam{} interpretability approach. To evaluate the efficacy of another interpretability technique if injected in \sei{}, we perform a second experiment. In Fig.~\ref{fig:SEI_grad} we report the result obtained with \emph{Gradients}, the other approach that together with \gradcam{} passes the ``sanity check'' of the interpretability approaches~\cite{adebayo2018sanity}. In practice, this approach amounts to remove from Eq.~\eqref{eq:gradcam} the feature map variable $F^{k,j}$. This brings to a sort of \cui{} curve where we summed the gradients associated with a given input without multiplying it by the features themselves.

As visible, \gradcam{} gives a better curve in terms of proportionality \wrt{} the final class accuracy, while \emph{Gradients} is growing as soon as the layer is deeper, making it hard to spot the best split point. This is due to the vanishing gradient effect: gradients are much higher when closer to the network's end. For this reason we prefer \gradcam{}, since the multiplication of the gradients by the values of the feature (close to 0) provides a more constrained range of \cui{} values.

\begin{figure}[t]
\begin{center}
\includegraphics[width=0.95\linewidth]{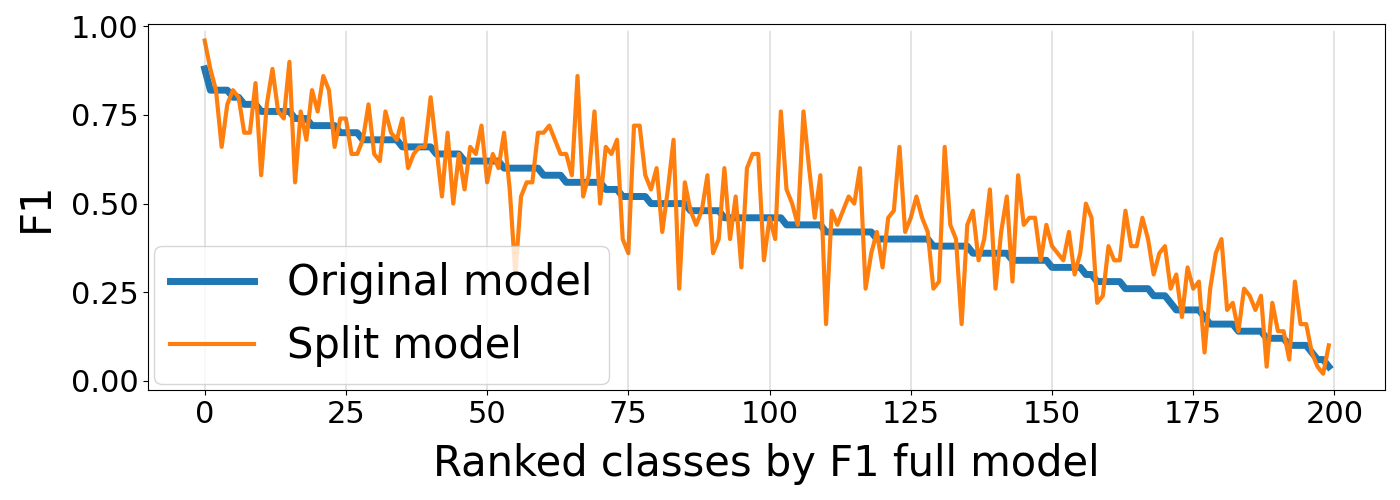}
\end{center}
\caption{The behavior of \sei{} on preserving the per-class performance before the split (original model, in blue) and after the split (split model, in orange). As expected, no dramatic changes in the single performance are visible, especially with the best and worse performing classes: the best performance does not become the worst after the split, and vice-versa.
\vspace{-1em}}
\label{fig:f1_exp}
\end{figure}
The third experiment investigates how the classification performance of a DNN does change, once it is split by our approach. In particular, we focus on the performance of every single class. As a split location, we select the layer $15$, where the \cui{} curve has the highest peak. %
After training the original VGG network, we rank the 200 classes in descending F1  order (the blue line of Fig.~\ref{fig:f1_exp}). As visible, the performance is unbalanced, since the F1 scores range from a few decimals to 87\%. On the (re-trained) split version we calculate new F1 scores, reporting them in the figure whilst keeping the previous class ranking to highlight possible variations in the performance (the orange line of Fig.~\ref{fig:f1_exp}). The plot is very interesting: actually, classes that were best or worst performing before the split do not change seriously their performance after the split, providing F1 scores which are still high/low. For classes with an average performance, the split shuffled a little the F1 scores. Overall, a certain degree of variability is expected, since \emph{(i)} the CUI curve indicating the optimal split point requires a summation over all the classes, so the per-class performance cannot be precisely preserved. \emph{(ii)} It seems that for specific classes the optimal splitting point may be different. We will focus on this aspect in the second part of the experiments.

\begin{figure}[t]
\centering
\begin{minipage}[t]{0.49\linewidth}
\centering
\includegraphics[width=\linewidth]{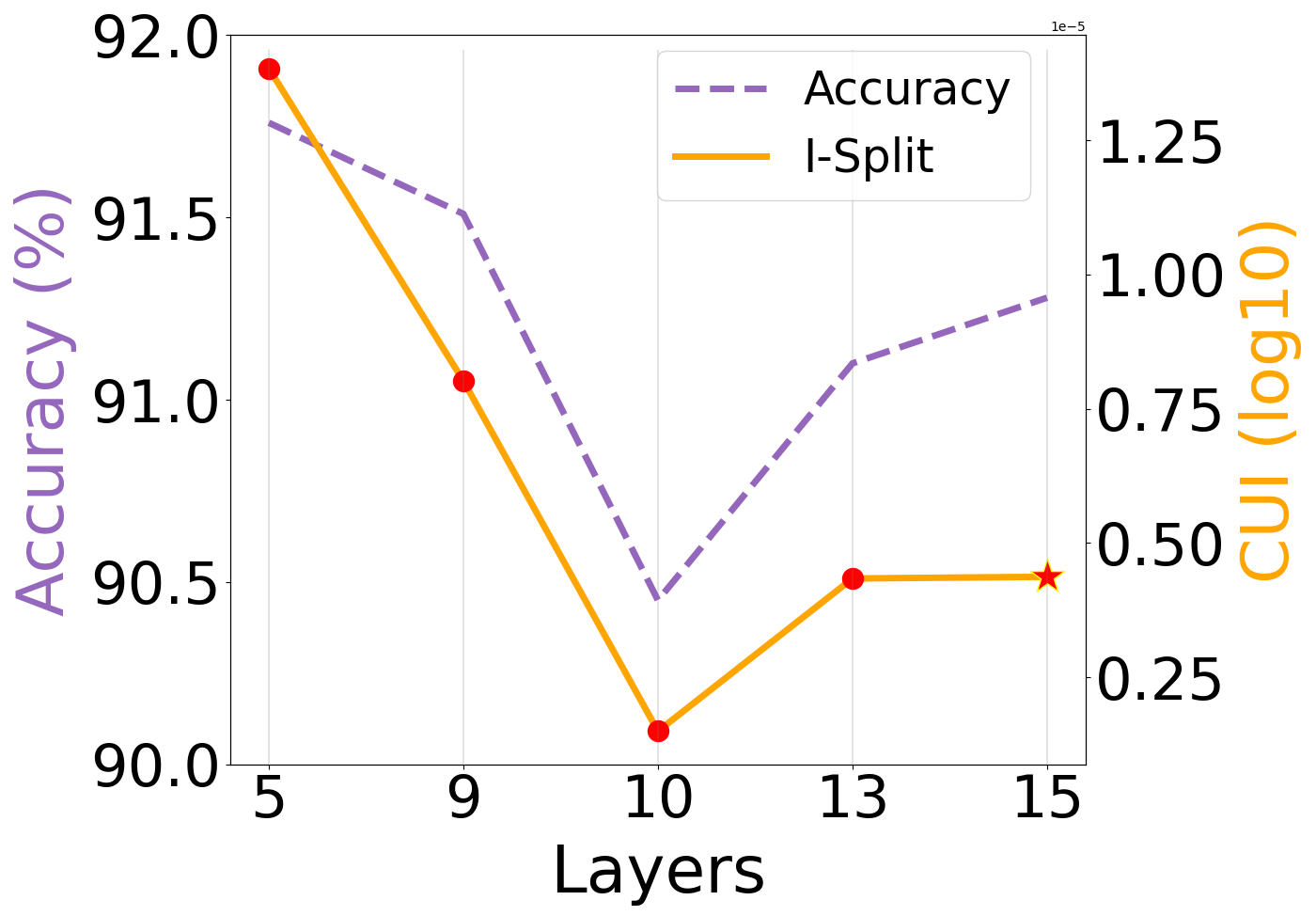}
\end{minipage}
\begin{minipage}[t]{0.49\linewidth}
\centering
\includegraphics[width=\linewidth]{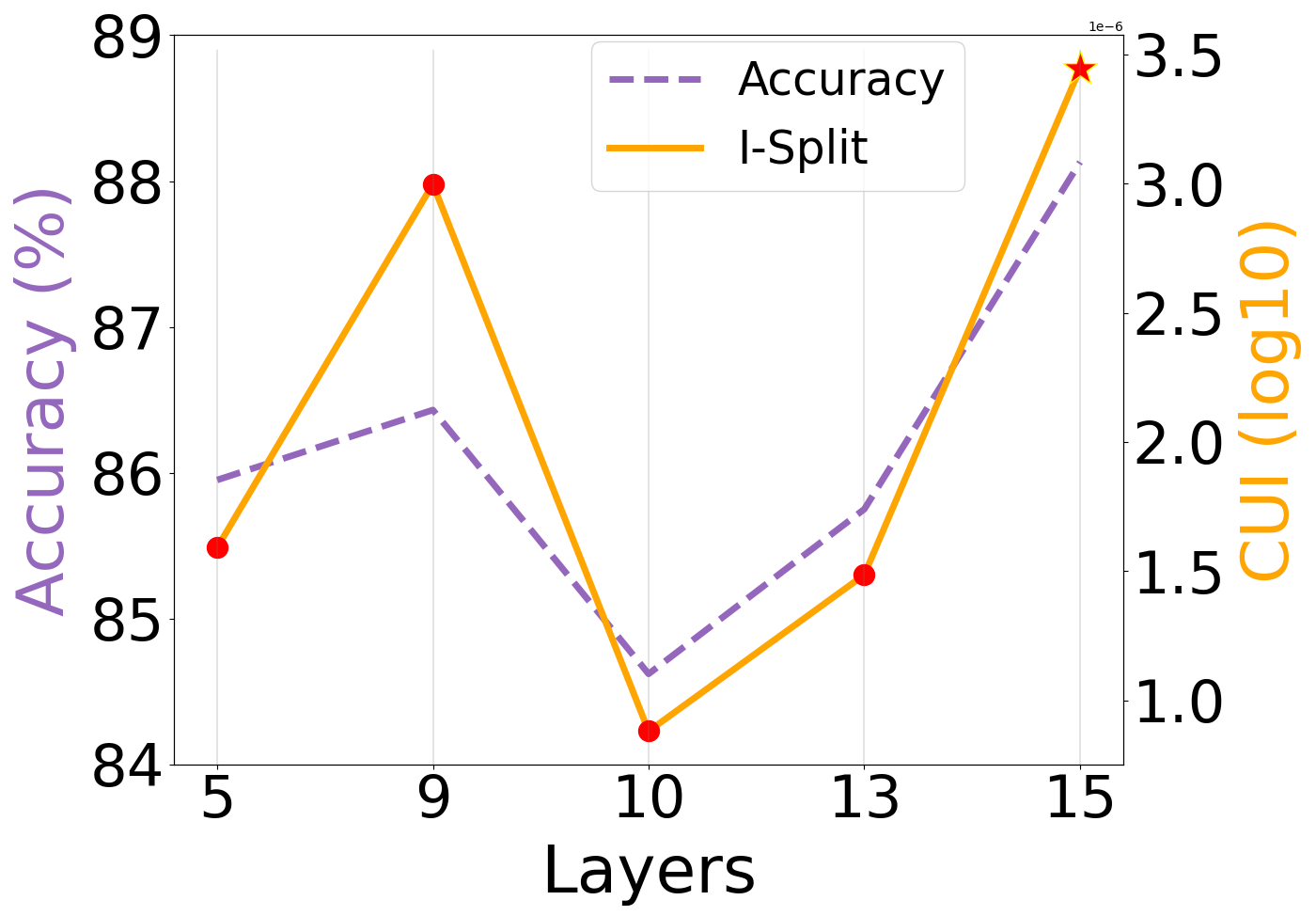}
\end{minipage}
\caption{The efficacy of \sei{} on different datasets. Both the figures show the CUI curve of \sei{} (solid yellow) and the accuracy when splitting between a particular layer, retraining, and testing (dashed purple). On the left, the notMNIST~\cite{notMNIST} dataset. On the right, the  Chest X-Ray Pneumonia~\cite{chest_xray} dataset. In both the cases, the \cui{} curve shows to be proportional to the accuracy.
\vspace{-.5em}}
\label{fig:notMNIST}
\end{figure}
The fourth and fifth experiments check how \sei{} will perform with datasets with fewer classes, possibly unbalanced in terms of cardinality. For this sake,  we apply our method to notMNIST~\cite{notMNIST} (10 classes) and Chest X-Ray Pneumonia~\cite{chest_xray} (2 classes, unbalanced). The notMNIST dataset is focused on the task of digit recognition with heterogeneous fonts and graphics, while Chest X-Ray focuses on classifying pneumonia cases from chest x-rays. Chest X-Ray is unbalanced (cases: 3883, controls: 1349), and this is a further challenge for the \sei{} robustness.

In both cases, the employed deep network is the VGG16. Results are reported in the form of \cui{} curves in Fig.~\ref{fig:notMNIST}, paired with the a posteriori accuracy. In both cases, the \cui{} curves are proportional to the accuracy. This further promotes \sei{} as a prognostic tool toward a split configuration which maximizes the classification accuracy.

\begin{figure}[t]
\begin{center}
\includegraphics[width=.95\linewidth]{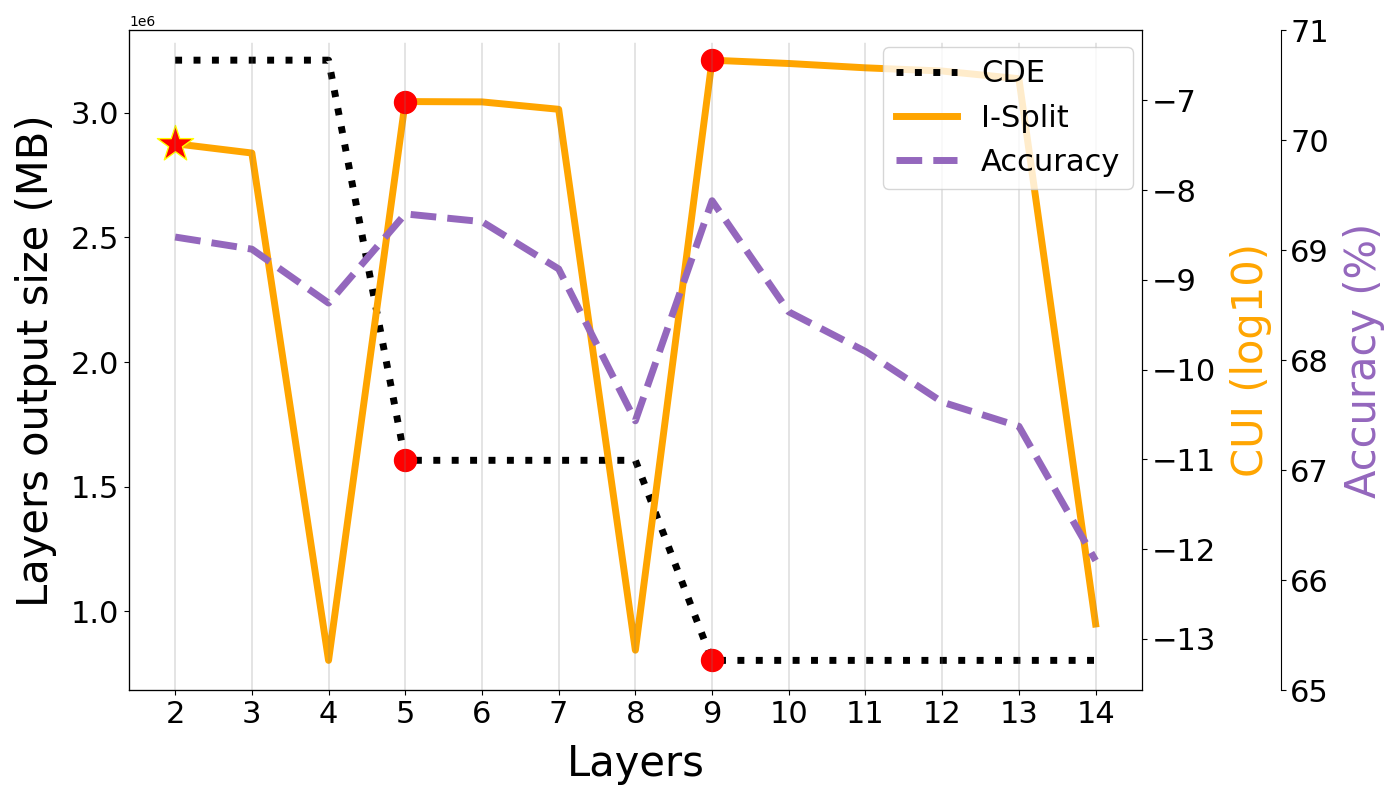}
\end{center}
\caption{Results of the CDE \cite{sbai2021cut}, \sei{}, and accuracy on the ResNet-50. In particular, the local minimum points of CDE correspond to the local maximum points of \sei{}, and the \cui{} curve shows to be once again proportional to the accuracy. The ``star'' markers show the extra points that our method is able to identify, and not identified by the CDE approach.
\vspace{-.5em}}
\label{fig:CDE_and_SEI_ResNet50}
\end{figure}

On the ResNet-50 architecture~\cite{he2016deep}, we show the \cui{} curve, the CDE~\cite{sbai2021cut} comparative approach, and the post-split accuracy curve (Fig.~\ref{fig:CDE_and_SEI_ResNet50}). As visible, the local maxima of the \cui{} curve (indicated by the red markers) indicate plausible split locations. Overall, the \cui{} curve follows the slope of the accuracy, showing even in this case nice predictive properties. Additionally, we can observe that the \cui{} is almost constant in layers with the same features dimensionality, but it is also relatively uniform between different blocks.

\subsection{Class-Dependent Splitting Point Selection} \label{sec:split2class}
\begin{figure}[t]
\begin{center}
\includegraphics[width=.9\linewidth]{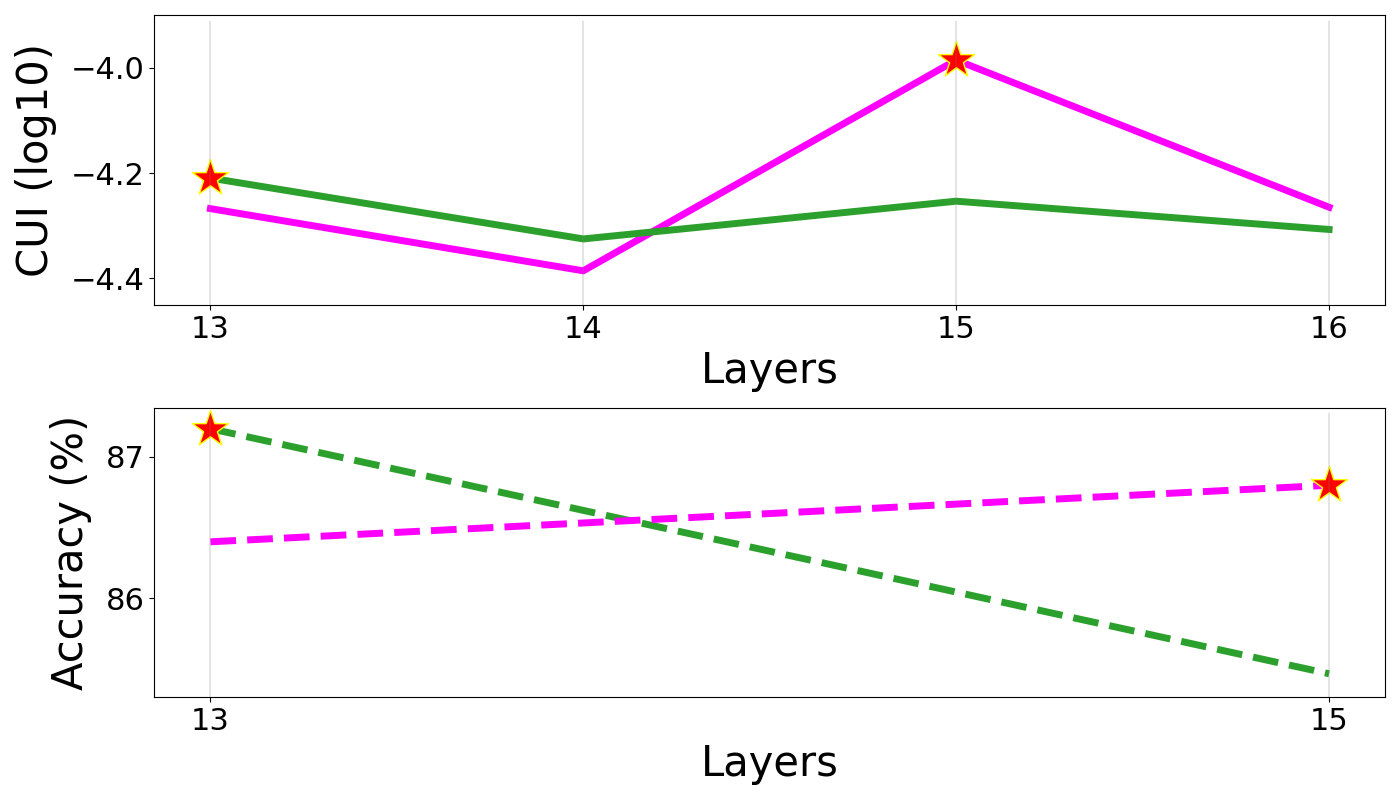}
\end{center}
\caption{The comparison of splitting results for different subsets of classes. Top: \cui{} curve computed on two different subsets of 15 classes of the VGG16 architecture. Bottom: classification accuracy for two models trained on the same subsets of 15 classes on layers $13$ and $15$. 
\vspace{-1.5em}}
\label{fig:SEI_and_accuracy_per_class_1and2_VGG16}
\end{figure}

In this work, we also claim that different subsets of classes can lead to different selections of the best splitting point, as guessed in the previous section. To validate this claim we analyzed the \sei{} curve for each class separately and we identified two different behaviors in the last portion of the network. In particular, when referring to \emph{block5}, we saw that most of the classes achieved a higher \cui{} at layer $15$, while 15 classes had a global maximum at layer $13$ (see Fig.~\ref{fig:SEI_and_accuracy_per_class_1and2_VGG16}-top). We investigated this phenomenon by generating two alternative models by splitting the network at both points and then retraining these models with two subsets of 15 classes, one for each different behavior. Results, shown in Fig.~\ref{fig:SEI_and_accuracy_per_class_1and2_VGG16}-bottom, demonstrate that the same shape of the \sei{} curve translates to the accuracy curve. 

\section{Conclusion}
\label{cha:conclusion}

In this paper we presented \sei{}, a new split computing method to identify the best splitting point in order to preserve higher classification accuracy in a DNN. %
Our approach exploits interpretability principles, specifically \gradcam{} approach, to extract an estimate of the importance of each layer of the network related to a classification task. The assumption is that feature maps with high importance values (dubbed \cui{}) are more suitable to be compressed and sent through a communication network. %
With respect to the state-of-the-art, \sei{} introduces major benefits: \emph{(i)} it is inherently faster than most of the competitors because it does not require retraining the model on more than one splitting point; \emph{(ii)} it is able to discriminate between layers with the same size, which are usually considered equivalent; and \emph{(iii)} it can provide different optimal splitting points according to priors on the classes the system is expected to deal with.
Future works will include further investigation of interpretability methods as a way to extract additional metrics to be used in the generation of the \sei{} curve, and the development of a bottleneck architecture specifically designed to preserve classification accuracy rather than preserving the input data. 

\section*{Acknowledgements}
The work has been partially supported by the Italian Ministry of Education, University and Research (MIUR) with the grant ``Dipartimenti di Eccellenza'' 2018-2022 and by Fondazione Cariverona with the grant ``Ricerca \& Sviluppo''.

\bibliographystyle{IEEEtran}
\bibliography{bibi}

\end{document}